# Optimized Hidden Markov Model based on Constrained Particle Swarm Optimization

Liu Chang [1,2,3], Yacine Ouzrout[1], Antoine Nongaillard[1],
Abdelaziz Bouras[1], Zhou Jiliu[2,3]

[1]DISP Laboratory, University Lumiere of Lyon 2, France
[2]Key Laboratory of Pattern Recognition and Intelligent Information Processing in Sichuan, China
[3]College of Information Science and Technology, Chengdu University, Chengdu, China

**As one of Bayesian analysis tools, Hidden Markov Model (HMM) has been used to in extensive applications. Most HMMs are solved by Baum-Welch algorithm (BWHMM) to predict the model parameters, which is difficult to find global optimal solutions. This paper proposes an optimized Hidden Markov Model with Particle Swarm Optimization (PSO) algorithm and so is called PSOHMM. In order to overcome the statistical constraints in HMM, the paper develops re-normalization and re-mapping mechanisms to ensure the constraints in HMM. The experiments have shown that PSOHMM can search better solution than BWHMM, and has faster convergence speed.**

## I. INTRODUCTION

As an efficient statistics tool, Hidden Markov Model (HMM) has been tested and proved in a wide range of applications. It is a powerful algorithm to estimate the model parameters. Therefore, to optimize the model parameter exactly is a crucial element in HMM.

To find optimal model parameters, traditional approach often use Baum-Welch (BW) algorithm to optimize model parameters based on expectation-maximization (EM) algorithm. However, BW algorithm often converge into local optimum.[1] re-estimates model parameters with reinforcement learning, but still cannot overcome local convergence problem. Recently, various intelligent evolution algorithms are introduced to optimize HMM and achieve good performance. [2] optimizes HMM by tabu search algorithm; [3] [4] proposes to training HMM structure with genetic algorithm(GA); [5] trains HMM by Particle Swarm Optimization(PSO) algorithm; [6, 7] make a comparison between PSO and GA for HMM training and demonstrate that hybrid algorithm based on PSO and BW is superior to BW algorithm and the hybrid algorithm based on GA and BW. Because model parameters need to satisfy statistical characteristic, the optimization of model parameters in HMM is a constraint problem. But these evolution algorithms usually combine evolution algorithms with HMM directly and leave these parameter constraints in HMM out of consideration.

In this paper, we employ Particle Swarm Optimization to search optimal model parameters of HMM (PSOHMM) to avoid local optimum in BW algorithm, solve the parameter constraints in HMM with remapping and re-normalization mechanism. The simulation experiments demonstrate PSOHMM achieves better optimization performance than BWHMM.

The remainder of this paper is organized as follows. In the next section, some basic knowledge of HMM is given. Section3 introduces Optimized Hidden Markov Model. In section4, we discuss the performance of the proposed algorithm. Finally, Section 5 concludes this paper and proposes future works.

## II. HIDDEN MARKOV MODEL

Given a set of $m$ observation states $V = \{v_1, \cdots, v_m\}$, HMM consists of a finite set of $n$ hidden states $S = \{s_1, \cdots, s_n\}$ with an associated probability distribution. It means that suppose that HMM regularly undergoes a state-change along a certain constant period of time according to a set of probabilities associated with its current state. So HMM is a probabilistic model with a collection of random variables $\{o_1, \cdots, o_t, q_1, \cdots, q_t\}$ where $O = \{o_1, \cdots, o_t\}$ is the sequence of observation states and $o_i \in V$, $i \in [1, t]$, and $Q = \{q_1, \cdots, q_t\}$ is the sequence of hidden states and $q_i \in S$, $i \in [1, t]$.

For simplicity, two conditional independence assumptions are given to make associated algorithms tractable as following:
(1).The $t$ th hidden variable, given the $(t-1)$ st hidden variable, is independent of previous variables, i.e. $P(q_{t+1} | q_t, \cdots, q_1) = P(q_{t+1} | q_t)$.
(2).The $t$ th observation depends only on the $t$ th state, i.e. $P(o_t | q_t, \cdots, q_1, o_1) = P(o_t | q_t)$.
The definition of HMM can be described as following:

**Definition 1** (Hidden Markov Model) Given a set of $m$ observation states $V$, HMM with a finite set of $n$ hidden states $S$ consist of a triple $\lambda = (\pi, A, B)$, where $\pi = \{\pi_i = P(q_1 = s_i)\}$ is the prior probabilities of $s_i$ being the first state of $Q$; $A = \{a_{ij}\}$ is the state transition probabilities matrix, $1 \leq i, j \leq n$, $a_{ij} = P\{q_{t+1} = s_j \mid q_t = s_i\}$ characterize the transition probability from hidden state $q_{t+1}$ into $q_t$, and $\sum_j a_{ij} = 1$; $B = \{b_{ij}\}$ is the emission probabilities matrix, $1 \leq i \leq m, 1 \leq j \leq n$, $b_{ij} = P\{o_t = i \mid q_t = j\}$ describes the relation between observation $o_t$ and hidden state $q_t$ at time t, and $\sum_j b_{ij} = 1$;

According to Definition 1, the triple $\lambda$ in HMM should satisfy the prior probabilities, the state transition probabilities matrix $A$ and the emission probabilities matrix $B$ should be non-negative and normalized, which also called non-negative and normalization constraint respectively.

Given an observation sequence $O = \{o_1, \cdots, o_t\}$, it can solve the following three basic problems[8]:

(1) Given $O$ and $\lambda$, how to compute the probability of observing sequence, i.e. $P(O \mid \lambda)$. The problem is called the evaluation problem.

(2) Given $O$ and $\lambda$, how to find a corresponding hidden states sequence that most probably generated an observed sequence. The problem is called the decoding problem.

(3) Given $O$, how to adjust the model parameter $\lambda$ to maximize $P(O \mid \lambda)$. The problem is called the learning problem.

According to the second and third problems, before the hidden states sequence that most probably generated observed sequence is found, it's necessary to compute optimal model parameters $\lambda$ to maximize the following objective function:

$$\max P(O \mid \lambda) \quad (1)$$

Given model parameter $\lambda$, the above probability of $O$ is obtained according to sum joint probability over all possible state sequence $Q$ as following:

$$P(O \mid \lambda) = \sum_Q P(O, Q \mid \lambda) P(Q \mid \lambda) \quad (2)$$

According to Bayesian theory, we can get:

$$P(O, Q \mid \lambda) = P(O \mid Q, \lambda) \cdot P(Q \mid \lambda) \quad (3)$$

Give a hidden state sequence, the likelihood of an observation sequence is equal to the product of the emission probabilities computed along the specific path:

$$P(O \mid Q, \lambda) = \prod_{t=1}^{T} P(o_t \mid q_t, \lambda) = b_{q_1, o_1} \cdot b_{q_2, o_2} \cdots b_{q_T, o_T} \quad (4)$$

Give model parameter $\lambda$, the probability of a state sequence $Q = \{q_1, \cdots, q_T\}$ can be computed by the product of the transition probabilities from one state to another state:

$$P(Q \mid \lambda) = \pi_{q_1} \cdot \prod_{t=1}^{T-1} a_{q_t, q_{t+1}} = \pi_{q_1} \cdot a_{q_1, q_2} \cdot a_{q_2, q_3} \cdots a_{q_{T-1}, q_T} \quad (5)$$

Let Eq. and (5) substitute into (3), then we can get:

$$P(O, Q \mid \lambda) = \sum_{q_1, \cdots, q_T} \pi_{q_1} \cdot a_{q_1, q_2} \cdot b_{q_1, o_1} \cdots a_{q_{T-1}, q_T} \cdot b_{q_T, o_T} \quad (6)$$

Therefore, the objective function in Eq. becomes:

$$f(\lambda) = \max \sum_{q_1, \cdots, q_T} \pi_{q_1} \cdot a_{q_1, q_2} \cdot b_{q_1, o_1} \cdots a_{q_{T-1}, q_T} \cdot b_{q_T, o_T} \quad (7)$$

Traditional method to solve Eq. is to use Baum-Welch algorithm, which also is called forward and backward algorithm. However, BW algorithm finds maximum-likelihood with iterative expectation-maximization (EM) algorithm and is sensitive to initial model parameters which are easy to trap into the local optimum.

III. OPTIMIZED HIDDEN MARKOV MODEL BASED ON CONSTRAINED PARTICLE SWARM OPTIMIZATION

Particle Swarm Optimization(PSO) algorithm is one kind of evolutionary method proposed by Kennedy and Eberhart [9]. It imitates social behavior of birds and fish to find global optimum for optimization problems and has been applied into various scientific fields [10, 11, 12, 13, 14].

For $n$ particles in PSO, a particle is a solution, which represents the position of a particle, the group solutions consist of a swarm. In order to find the best position of the particle, i.e. the optimal solution, PSO constantly updates the positions of all particles with a velocity vector in the swarm in iterative manner until the termination condition met. For example, each particle $i$ is represented by a $D$ dimensional position vector $x_i(t)$ and has a corresponding velocity vector $v_i(t)$. During each iteration $t$, the best position of a particle and the swarm $pbest$, $gbest$ that result in the best value of fitness function are recorded, which are called the particles' personal experience and the social knowledge respectively, then the velocity vector is updated using the following rule:

$$v_i(t) = \omega v_i(t-1) + c_1 \xi_1 (pbest - x_i(t-1)) + c_2 \xi_2 (gbest - x_i(t-1)) \quad (8)$$

The parameter $\omega$ is called inertia weight to scale the previous time step velocity; $c_1$ and $c_2$ are scalar factor to control the influence of the personal experience and social knowledge; $\xi_1$ and $\xi_2$ represent the random number that satisfy a uniform distribution and $\xi_1, \xi_2 \in [0,1]$. The new position of the particle is updated:

$$x_i(t) = x_i(t-1) + v_i(t) \quad (9)$$

If the best position of the swarm $gbest$ is searched from the positions of all particles, then the PSO algorithm is called global best PSO; if $gbest$ is searched only from the neighborhood $N_i(t)$ of the particle $x_i$, i.e. $|N_i(t)|<n$, then the PSO algorithm is called local best PSO.

The fitness functions $f(x)$ derived from Eq. to find optimal model parameters $\lambda$ to maximize the probability of observation sequence $O$:

$$f(\lambda) = P(O|\lambda) \quad (10)$$

Since all model parameters $\lambda$ represent the probability, so $\lambda \in [0,1]$. However, it's easy to find that the position update rule in Eq.(8) cannot only guarantee model parameters are within $[0,1]$, but it also cannot satisfy the normalization constraint of model parameters. The reason is that PSO was originally proposed to handle unconstrained optimization. So it's necessary to modify PSO algorithm to cope with constrained optimization.

To cope with the constraints in HMM, this paper employs two different methods to guarantee the interval and normalization constraints respectively.

Concerning the problem that model parameters exceed number interval, we employ re-mapping method to adjust these parameters. Let $U_b = 1$ denotes upper bound, $L_b = 0$ denotes lower bound, $\psi$ denotes these model parameters exceeded number interval. The re-mapping method repairs $\psi$ considering the following four cases:

$$\psi = \begin{cases} U_b - \xi(\psi - U_b) & if\ \psi > U_b \\ L_b + \xi(L_b - \psi) & if\ \psi < L_b \end{cases} \quad (11)$$

Meanwhile, velocity vector $v_i(t)$ is also updated:

$$v_i(t) = x_i(t) - x_i(t-1) \quad (12)$$

Furthermore, in order to ensure that transition matrix and emission matrix satisfy the normalization constraints, the following re-normalization method is employed:

$$\pi_i = \frac{\pi_i}{\sum_{i=1}^{N} \pi_i} \quad (13)$$

$$\rho_{ij} = \frac{\rho_{ij}}{\sum_{j=1}^{N} \rho_{ij}} \quad (14)$$

Where $\rho_{ij}$ denotes $a_{ij}$ and $b_{ij}$.

## IV. SIMULATION EXPERIMENTS

In this section, two simulation experiments are conducted to validate the performance of the algorithm proposed in this paper. Firstly, we demonstrate that optimized HMM based on PSO( called PSOHMM) can search better solution than HMM solved BW algorithm(called BWHMM) and analyze the reason.

In the first experiments, two group datasets are generated randomly. To improve the efficiency of computation, we select a small number as the number of observation states and let it equal 5. In some e-commerce environments, the feedback often contains multi-dimensional rate scores, such as the rate to the quality of products, the quality of service and the delivery time. In order to imitate the rating score, we generate five different 1-dimension observation sequences with the length of 100 in the first group and generate five different 2-dimension observation sequences with the length of 100.

For each dataset, the model parameters of HMMs are trained using these observation sequences respectively and the number of hidden state is 2. At last, we can get five different HMMs. As shown in Section 2, the optimal model parameter should maximize the probability of observation sequence. So the log likelihoods of these observation sequences with these optimized HMMs are computed, i.e. $\log P(O|\lambda)$ to compare the optimization ability between PSOHMM and BWHMM as shown in Fig. 1 and Fig. 2. For PSOHMM, the number of particle is 25 and the iteration number is 10. For BWHMM, the iteration number is 50.

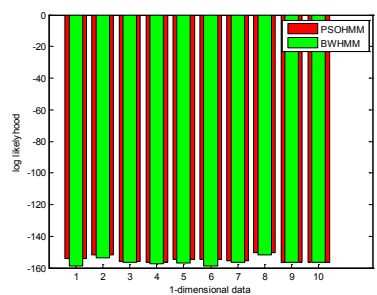

Fig.1 The log likelihoods of 1-dimensional data

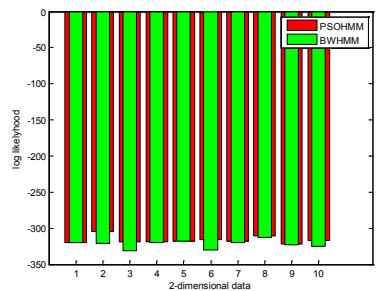

Fig. 2 The log likelihoods of 2-dimensional data

From Fig.1 and Fig.2, we can find easily that PSOHMM always achieves larger log likelihoods of observation sequences than BWHMM for both two data sets with one dimension and two dimension data. It means BWHMM is easy to trap into local optimum, but PSOHMM has the capability to find global optimal solution and can be applied into any dimension observation sequence. Actually, in PSOHMM, the current best position of a particle *pbest* represents a local optimum, *gbest* represents the best solution of all local optimum found in current iteration, as shown in Fig.3. In each iteration, the position is update with velocity vector, which means to search a better solution than previous local optimum within the neighborhood of previous position and make *gbest* move toward to global optimal solution continuously. Moreover, PSOHMM employs some particles to search optimal solution simultaneously and have larger probability to find global optimal solution, So PSOHMM is superior to BWHMM to search model parameters.

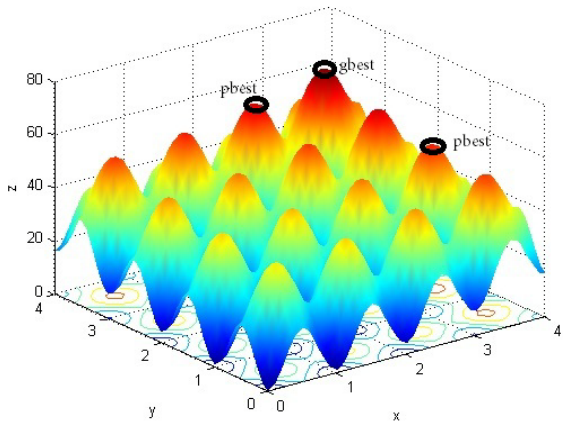

Fig.3 The global and local solution in Rastrigin's Function

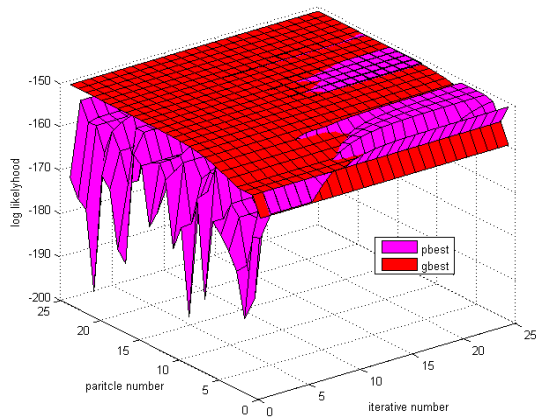

Fig.4 The convergence of *pbest* and *gbest* with iterative number

Fig.4 exactly verifies the above analysis. It shows that the best position of every particle *pbest* always converges to *gbest* with the increase of iteration number.

In order to demonstrate the convergence of PSOHMM, Fig.5 compares the convergence performance of PSOHMM and BWHMM with different iteration number. We can find that PSOHMM converge gradually after one iteration, but BWHMM begin to converge after 10 iterations. So PSOHMM has better convergence performance than BWHMM and more stable than BWHMM.

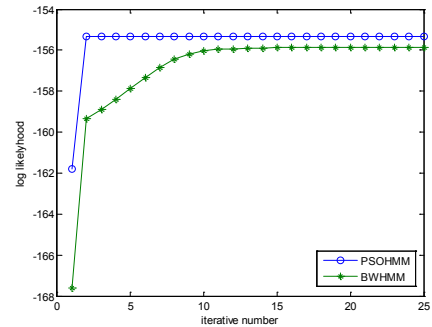

Fig.5 The convergence of log likelihoods with different iterative number

V. CONCLUSION

This paper proposes optimization HMM based on Particle Swarm Optimization. The proposed algorithm takes full use of the global searching capability of PSO and avoids BW algorithm trap into the local optimum. In addition, this paper employs re-mapping and re-normalized methods to guarantee the interval and normalized constraints in HMM. The simulation experiments have demonstrated that PSOHMM is superior to BW algorithm to search optimal model parameters and more stable than BW algorithm.

**Chang Liu** received the B.S. and M.S. degrees from School of Computer Science, Sichuan Normal University, Chengdu, China, in 2004 and 2007, respectively. She is currently pursuing the Ph.D. degree in the college of computer, Sichuan University.
Her research interests focus on machine learning, computer vision, pattern recognition and image processing.